\documentclass{article}

\usepackage[utf8]{inputenc} 
\usepackage[T1]{fontenc} 
\usepackage{hyperref} 
\usepackage{url} 
\usepackage{booktabs} 
\usepackage{amsfonts} 
\usepackage{nicefrac} 
\usepackage{microtype} 
\usepackage{graphicx} 
\usepackage{array}
\usepackage{amsmath}
\usepackage{placeins}
\usepackage{subcaption}
\usepackage{caption}
\usepackage[square,sort,comma,numbers]{natbib}
\usepackage{fancyhdr}
\usepackage{tabularx}

\usepackage[
a4paper,
textheight=9in,
textwidth=6in,
top=1in
]{geometry}
\renewenvironment{abstract}%
{%
\vskip 0.075in%
\centerline%
{\large\bf Abstract}%
\vspace{0.5ex}%
\begin{quote}%
}
{
\par%
\end{quote}%
\vskip 1ex%
}

\date{}

\title{Training Deep Spiking Neural Networks using Backpropagation}
\author{
Jun Haeng Lee$^*$$^\dag$, Tobi Delbruck$^\dag$, Michael Pfeiffer$^\dag$ \\
$^*$Samsung Advanced Institute of Technology, Samsung Electronics \\
\texttt{junhaeng2.lee@samsung.com} \\
$^\dag$Institute of Neuroinformatics, University of Zurich and ETH Zurich \\
\texttt{\{tobi, pfeiffer\}@ini.uzh.ch} \\
}
\begin{document}
\maketitle

\begin{abstract}
  Deep spiking neural networks (SNNs) hold great potential for improving the latency 
  and energy efficiency of deep neural networks through event-based computation. 
  However, training such networks is difficult due to the non-differentiable nature of 
  asynchronous spike events. In this paper, we introduce a novel technique, which 
  treats the membrane potentials of spiking neurons as differentiable signals,
  where discontinuities at spike times are only considered as noise. This enables 
  an error backpropagation mechanism for deep SNNs,
  which works directly on spike signals and membrane potentials.
  Thus, compared with previous methods relying 
  on indirect training and conversion, our technique has the potential to capture the 
  statics of spikes more precisely. Our novel framework outperforms all previously 
  reported results for SNNs on the permutation invariant MNIST benchmark, as well as
  the N-MNIST benchmark recorded with event-based vision sensors.
\end{abstract}

\section{Introduction} 
\label{introduction}

Deep learning is achieving outstanding results in various machine learning tasks
\cite{he2015deep, lecun2015deep}, but for applications that 
require real-time interaction with the real environment, the repeated and often redundant 
update of large numbers of units becomes a bottleneck for efficiency.  An alternative 
has been proposed in the form of spiking neural networks (SNNs), a major research 
topic in theoretical neuroscience and neuromorphic engineering. SNNs exploit 
event-based, data-driven updates to gain efficiency, especially if 
they are combined with inputs from event-based sensors, which reduce redundant 
information based on asynchronous event processing \cite{camunas2012event,merolla2014million,oconnor2013real}. Even though in theory 
\cite{maass2004computational} SNNs have been shown to be as computationally powerful as conventional artificial 
neural networks (ANNs, this term will be used to describe conventional deep neural 
networks in contrast with SNNs), practically SNNs have not quite reached the same accuracy levels of ANNs
in traditional machine learning tasks. A major reason for this is the lack 
of adequate training algorithms for deep SNNs, since spike signals are not differentiable,
but differentiable activation functions are fundamental for using error backpropagation.
A recently proposed solution is to use different data representations between training and processing, 
i.e. training a conventional ANN and developing conversion algorithms that transfer the 
weights into equivalent deep SNNs \cite{diehl2015fast,esser2015backpropagation,hunsberger2015spiking,oconnor2013real}. 
However, in these methods, details of statistics in spike trains that go beyond mean rates, such as required for processing event-based sensor data
cannot be precisely represented by the signals used for training. 
It is therefore desirable to devise learning rules operating directly on spike trains, but so far it has only been
possible to train single layers, and use unsupervised learning rules, which leads to a deterioration
of accuracy \cite{diehl2015unsupervised,masquelier2007unsupervised,neftci2014event}. 
An alternative approach has recently been introduced by \cite{oconnor2016deep}, in which a
SNN learns from spikes, but requires keeping statistics for computing stochastic gradient descent (SGD) updates
in order to approximate a conventional ANN.
In this paper we introduce a novel supervised learning 
technique, which can train general forms of deep SNNs directly from spike signals.
This includes SNNs with leaky membrane potential and spiking 
winner-takes-all (WTA) circuits.
The key idea of our approach is to generate a continuous and differentiable signal on which SGD can work, 
using low-pass filtered spiking signals added onto the membrane potential
and treating abrupt changes of the membrane potential as noise during error backpropagation.
Additional techniques are presented that address 
particular challenges of SNN training: 
spiking neurons typically require large thresholds to achieve stability and reasonable firing rates,
but this may result in many “dead” neurons, which 
do not participate in the optimization during training. Novel regularization and normalization techniques
are presented, which contribute to stable and balanced learning. 
Our techniques lay the foundations for closing the performance gap between SNNs and ANNs, and promote
their use for practical applications.

\section{Related Work} 
\label{related_work}

Gradient descent methods for SNNs have not been deeply investigated because of the 
non-differentiable nature of spikes. The most successful approaches to date have used 
indirect methods, such as training a network in the continuous rate domain and converting
it into a spiking version. 
O’Connor et al. pioneered this area by training a spiking deep 
belief network (DBN) based on the Siegert event-rate approximation model \cite{oconnor2013real}, but only 
reached accuracies around $94.09\%$ for the MNIST hand written digit classification task. 
Hunsberger and Eliasmith used the softened rate model for leaky integrate and fire (LIF) 
neurons \cite{hunsberger2015spiking}, training an ANN with the rate model and converting it into a SNN consisting 
of LIF neurons. With the help of pre-training based on denoising autoencoders they 
achieved $98.6\%$ in the permutation-invariant (PI) MNIST task. Diehl et al. \cite{diehl2015fast} trained deep neural 
networks with conventional deep learning techniques and additional constraints necessary 
for conversion to SNNs. After the training units were converted into spiking neurons 
and the performance was optimized by normalization of weight parameters, yielding
$98.64\%$ accuracy in the PI MNIST task. Esser et al. \cite{esser2015backpropagation} used a differentiable 
probabilistic spiking neuron model for training and statistically sampled the trained 
network for deployment. In all of these methods, training was performed indirectly 
using continuous signals, which 
may not capture important statistics of spikes generated by sensors used during processing time.
Even though SNNs are optimally suited for processing signals 
from event-based sensors such as the Dynamic Vision Sensor (DVS) \cite{lichtsteiner2008dvs}, the previous 
SNN training models require to get rid of time information and generate image frames from 
the event streams. Instead, we use the 
same signal format for training and processing deep SNNs, and can thus train SNNs directly
on spatio-temporal event streams. This is demonstrated on the neuromorphic N-MNIST benchmark dataset \cite{orchard2015converting},
outperforming all previous attempts that ignored spike timing.

\section{Spiking Neural Networks} 
\label{snn}

In this article we study fully connected SNNs with multiple hidden layers. Let $M$ and $N$ be
the number of synapses of a neuron and the number of neurons in a layer, respectively. 
On the other hand, $m$ and $n$ are the number of active synapses (i.e. synapses receiving spike inputs) 
of a neuron and the number of active neurons (sending spike outputs) in a layer. We will 
also use the simplified form of indices for active synapses and neurons throughout the 
paper as
\begin{center}
  \emph{Active synapses:} \{$v_1, \cdots, v_m$\}$\rightarrow$\{$1, \cdots, m$\}, 
  \emph{Active neurons:} \{$u_1, \cdots, u_n$\}$\rightarrow$\{$1, \cdots, n$\}
\end{center}

Thus, if an index $i$, $j$, or $k$ is used for a synapse over [1, $m$] or 
a neuron over [1, $n$] (e.g. in (\ref{eq:mp_full})), it actually represents an index of an active 
synapse ($v_i$) or an active neuron ($u_j$). 

\subsection{Leaky Integrate-and-Fire (LIF) Neuron} 
\label{lifn}

The LIF neuron is one of the simplest models used for describing dynamics of spiking 
neurons \cite{gerstner2002spiking}. Since the states of LIF neurons can be updated 
asynchronously based on the timing of input events, it is a very efficient model in terms 
of computational cost. For a given input spike the membrane potential of a LIF 
neuron can be updated as
\begin{equation}
  \label{eq:lifmp}
  V_{mp} (t_p)=V_{mp}(t_{p-1})e^{\frac{t_{p-1} - t_p}{\tau_{mp}}} + w_i^{(p)}w_{dyn},
\end{equation}
where $V_{mp}$ is the membrane potential, $\tau_{mp}$ is the membrane time constant, 
$t_p$ and $t_{p-1}$ are the present and previous input spike times,  
$w_i^{(p)}$ is the synaptic weight of the $i$-th synapse (through which the
present $p$-th input spike arrives). $w_{dyn}$ is a dynamic weight controlling the 
refractory period, defined as $w_{dyn} = w_{d0}+(\Delta_t/T_{ref})^2$ if $\Delta_t < T_{ref}$ and $w_{dyn}<1$, and $w_{dyn}=1$ 
otherwise. $T_{ref}$ is  the refractory period, $w_{d0}$ is the initial value (usually $0$), 
and $\Delta_t = t_{out} - t_p$, where $t_{out}$ is the time of the latest output spike produced 
by the neuron.
Thus, the effect of input spikes on $V_{mp}$ is suppressed for a 
short period of time $T_{ref}$ after an output spike. $w_{dyn}$ recovers quadratically 
to $1$ after the output spike at $t_{out}$. Since $w_{dyn}$ is applied to all synapses 
identically, it is different from short-term plasticity, which is a synapse specific mechanism. 
When $V_{mp}$ crosses the threshold value $V_{th}$, the LIF neuron generates an 
output spike and $V_{mp}$ is decreased by a fixed amount proportional to the threshold:
\begin{equation}
  \label{eq:mpreset}
  V_{mp} (t_p^+)=V_{mp}(t_p) - \gamma V_{th},
\end{equation}
where $\gamma$ is the membrane potential reset factor and $t_p^+$ is time right after the reset. 
We used $\gamma = 1$ for 
all the results in this paper. The valid range of the membrane potential is limited to 
[$-V_{th}$, $V_{th}$]. Since the upper limit is guaranteed by (\ref{eq:mpreset}), the 
membrane potential is clipped to $-V_{th}$ when it falls below this value. This strategy 
helps balancing the participation of neurons during training. We will revisit this issue 
when we introduce threshold regularization in Section \ref{th_regularization}.

\subsection{Winner-Take-All (WTA) Circuit} 
\label{wta}

We found that the accuracy of SNNs could be improved by introducing a competitive 
recurrent architecture called WTA circuit in certain layers. In a WTA 
circuit, multiple neurons form a group with lateral inhibitory connections. 
Thus, as soon as any neuron produces an output spike, it inhibits all other neurons in 
the circuit and prevents them from spiking \cite{rozell2008sparse}. In this work, all lateral connections 
in a WTA circuit have the same strength, which reduces memory and computational 
costs for implementing them. The amount of lateral inhibition applied to the membrane 
potential is designed to be proportional to the inhibited neuron’s membrane potential 
threshold (see (\ref{eq:mp_full}) in Section \ref{tfunction}). With this scheme, lateral 
connections inhibit neurons having small $V_{th}$ weakly and those having large 
$V_{th}$ strongly. This improves the balance of activities among neurons 
during training. As shown in Results, WTA competition in the SNN led to remarkable improvements, especially 
in networks with a single hidden layer. 
The WTA circuit also improves the stability and speed of training. 

\section{Using Backpropagation in SNNs}
\label{backprop}

We now derive the transfer function for spiking neurons in WTA configuration 
and the SNN backpropagation equations. We also introduce simple methods to initialize 
parameters and normalize backpropagating errors to address vanishing or exploding 
gradients, and to stabilize training. 

\subsection{Transfer function and derivatives}
\label{tfunction}

From the event-based update in (\ref{eq:lifmp}), the accumulated effects of the $k$-th synapse
onto the membrane potential (normalized by synaptic weight) and the membrane potential 
reset in (\ref{eq:mpreset}) (normalized by $\gamma V_{th}$) at time $t$ can be derived as
\begin{equation}
  \label{eq:activity}
  x_k(t)=\sum_{p}{\exp \left(\frac{t_p - t}{\tau_{mp}}\right)}, 
  \quad a_i(t)=\sum_{q}{\exp \left(\frac{t_q - t}{\tau_{mp}}\right)},
\end{equation}
where the sum is over all input spike times $t_p<t$ of the synapse for $x_k$, and the output spike times $t_q<t$ for $a_i$. 
The accumulated effects of lateral inhibitory signals in WTA circuits 
can be expressed analogously to (\ref{eq:activity}). 
Ignoring the effect of refractory periods for now, this means that 
the membrane potential of the $i$-th active neuron in a WTA circuit can be written as
\begin{equation}
  \label{eq:mp_full}
  V_{mp,i} (t)=\sum_{k=1}^{m}{w_{ik}x_k(t)} - \gamma V_{th,i}a_i(t) 
                    + \sigma V_{th,i}\sum_{j=1, j\neq i}^{n} {\kappa_{ij}a_j(t)}. 
\end{equation}
The terms on the right side represent the input, membrane potential resets, and
lateral inhibition, respectively. 
$x_k$ denotes the effect of the $k$-th active input neuron, and $a_i$ 
the effect induced by output activity of the $i$-th active neuron, as defined in (\ref{eq:activity}).
$\kappa_{ij}$  is the strength of lateral inhibition ($-1 \leq \kappa_{ij} < 0$) from the $j$-th active 
neuron to the $i$-th active neuron, and $\sigma$ is the expected efficacy of lateral 
inhibition. $\sigma$ should be smaller than $1$, since lateral inhibitions can affect the 
membrane potential only down to its lower bound (i.e. $-V_{th}$). 
We found a value of $\sigma \approx 0.5$ to work well in practice.
Eq. (\ref{eq:mp_full}) reveals the relationship between 
inputs and outputs of spiking neurons which is not clearly shown in (\ref{eq:lifmp}) 
and (\ref{eq:mpreset}). Since the output ($a_i$) of the current layer becomes the 
input ($x_k$) of the next layer if all the neurons have same $\tau_{mp}$, (\ref{eq:mp_full}) 
provides the basis for backpropagation. Differentiation is not defined in (\ref{eq:activity}) at 
the moment of each spike because of a step jump. 
However, we can regard these jumps 
as noise while treating (\ref{eq:activity}) and (\ref{eq:mp_full}) as differentiable continuous 
signals to derive derivatives for backpropagation. In previous works  \cite{diehl2015fast,esser2015backpropagation,hunsberger2015spiking,oconnor2013real}, continuous variables were introduced 
as a surrogate for $x_k$ and $a_i$ in (\ref{eq:mp_full}) for backpropagation. In this 
work, however, we directly use the contribution of spike signals to the membrane 
potential as defined in (\ref{eq:activity}). 
Thus, the real statistics of spike signals, including temporal effects such as synchrony between inputs,
can influence the training process.
Ignoring the step jumps caused by spikes in the calculation of gradients might of course introduce
errors, but we found in practice that this has very little influence on SNN training.
A potential explanation is that by regarding the signals in (\ref{eq:activity}) as continuous signals, but corrupted by noise at the times of spikes,
this can have a similar positive effect as the widely used approach of noise injection during
training, which can improve the generalization capability of neural networks \cite{vincent2008extracting}.
In the case of SNNs, several papers have used the trick of treating spike-induced 
abrupt changes as noise for gradient descent optimization \cite{bengio2015objective,hunsberger2015spiking}. 
However, in these cases the model added Gaussian random noise instead of spike-induced pertubations.
In this work, we directly use the actual contribution of spike signals to the membrane potential 
as described in (\ref{eq:activity}) for training SNNs. 
Here we show that this approach works well for learning in SNNs where information is encoded in spike rates,
but importantly, the presented framework also provides the basis for utilizing specific 
spatio-temporal codes, which we demonstrate on a task using directly inputs from event-based 
sensors.

For the backpropagation equations we need to obtain the transfer functions of LIF neurons in the WTA circuit.
For this we set the residual $V_{mp}$ term in the left side of (\ref{eq:mp_full}) to zero (since it is not 
relevant to the transfer function), resulting in the transfer function
\begin{equation}
  \label{eq:transfer_func}
  a_i \approx \frac{s_i}{\gamma V_{th,i}} 
                  + \frac{\sigma \sum_{j = 1, j \neq i}^{n}{\kappa_{ij}a_j}}{\gamma},
  \text{  where  } s_i = \sum_{k=1}^{m}{w_{ik}x_k}.
\end{equation}
Refractory periods are not considered here since the activity of neurons in SNNs is rarely 
dominated by refractory periods in a normal operating regime. For example, we used a 
refractory period of $1$ ms and the event rates of individual neurons were kept within a 
few tens of events per second (eps). Eq. (\ref{eq:transfer_func}) is consistent with (4.9) in \cite{gerstner2002spiking} without WTA 
terms. It can also be simplified to a spiking version of a rectified-linear unit by introducing a unit threshold 
and non-leaky membrane potential as in \cite{oconnor2016deep}.   
Directly differentiating (\ref{eq:transfer_func}) yields the backpropagation equations 
\begin{equation}
  \label{eq:da_ds}
  \frac{\partial a_i}{\partial s_i} \approx \frac{1}{\gamma V_{th,i}},
  \frac{\partial a_i}{\partial w_{ik}} \approx \frac{\partial a_i}{\partial s_i}x_k,
  \frac{\partial a_i}{\partial V_{th,i}} \approx \frac{\partial a_i}{\partial s_i}(-\gamma a_i + \sigma \sum_{j \neq i}^{n} {\kappa_{ij} a_j}),
  \frac{\partial a_i}{\partial \kappa_{ih}} \approx \frac{\partial a_i}{\partial s_i}(\sigma V_{th,i}a_h),
\end{equation}
\begin{equation}
  \label{eq:da_dx}
  \begin{bmatrix}
    \frac{\partial a_1}{\partial x_k} \\
    \vdots \\
    \frac{\partial a_1}{\partial x_k}
  \end{bmatrix}
  \approx \frac{1}{\sigma} 
  \begin{bmatrix}
    q & \cdots & -\kappa_{1n} \\
    \vdots & \ddots & \vdots \\
    -\kappa_{n1} & \cdots & q
  \end{bmatrix}^{-1}
  \begin{bmatrix}
    \frac{w_{1k}}{V_{th,1}} \\
    \vdots \\
    \frac{w_{nk}}{V_{th,n}}
  \end{bmatrix}
\end{equation}
where $q=\gamma/\sigma$. When all the lateral inhibitory connections have the same strength 
($\kappa_{ij} = \mu, \forall i, j$) and are not learned, $\partial a_i/\partial \kappa_{ih}$ 
is not necessary and (\ref{eq:da_dx}) can be simplified to
\begin{equation}
  \label{eq:da_dx2}
  \frac{\partial a_i}{\partial x_k} \approx \frac{\partial a_i}{\partial s_i} \frac{\gamma}{(\gamma-\mu\sigma)}
  \left(
    w_{ik} - \frac{\mu\sigma V_{th,i}}{\gamma+\mu\sigma (n-1)} \sum_{j=1}^{n}{\frac{w_{jk}}{V_{th,j}}}
  \right).
\end{equation}
We consider only the first-order effect of the lateral connections in the derivation of gradients.  
Higher-order terms propagating back through multiple lateral connections are neglected 
for simplicity. This is mainly because all the lateral connections considered here are 
inhibitory. For inhibitory lateral connections, the effect of small parameter changes
decays rapidly with connection distance. Thus, first-order approximation 
saves a lot of computational cost without loss of accuracy.

\subsection{Initialization and Error Normalization}
\label{normalization}

Good initialization of weight parameters in supervised learning is critical to handle the 
exploding or vanishing gradients problem in deep neural networks \cite{glorot2010understanding,he2015delving}. The basic 
idea behind those methods is to maintain the balance of forward activations and backward 
propagating errors among layers. Recently, the batch normalization technique has been 
proposed to make sure that such balance is maintained through the whole training 
process \cite{ioffe2015batch}. However, normalization of activities as in the batch normalization scheme is
difficult for SNNs, because there is no efficient method for amplifying event rates. The initialization 
methods proposed in \cite{glorot2010understanding,he2015delving} are not appropriate for SNNs either, because SNNs have 
positive thresholds that are usually much larger than individual weight values. In this work, 
we propose simple methods for initializing parameters and normalizing backprop errors 
for training deep SNNs. Even though the proposed technique does not guarantee the 
balance of forward activations, it is effective for addressing the exploding and vanishing 
gradients problems.

The weight and threshold parameters of neurons in the $l$-th layer are initialized as
\begin{equation}
\label{eq:initialization}
  w^{(l)} \sim U\left[ -\sqrt{3/M^{(l)}},  \sqrt{3/M^{(l)}}\right], \quad
  V_{th}^{(l)}=\alpha\sqrt{3/M^{(l)}}, \quad \alpha > 1,
\end{equation}
where $U[-a, a]$ is the uniform distribution in the interval $(-a, a)$, $M^{(l)}$ is the 
number of synapses of each neuron, and $\alpha$ is a constant. $\alpha$ should be 
large enough to stabilize spiking neurons, but small enough to make the neurons 
respond to the inputs through multiple layers. We used values between 3 and 10 for 
$\alpha$. The weights initialized by (\ref{eq:initialization}) satisfy the following condition: 
\begin{equation}
\label{eq:weight_condition}
  E\left[\sum_i^{M^{(l)}}(w_{ji}^{(l)})^2\right] = 1 \quad \text{or} \quad E\left[ (w_{ji}^{(l)})^2\right]=\frac{1}{M^{(l)}}. 
\end{equation}
This condition is used for backprop error normalization in the next paragraph. 
In addition, to ensure stability, the weight parameters are regularized by decaying them 
so that they do not deviate too much from (\ref{eq:weight_condition}) throughout training. 
We will discuss this in detail in Section 5.1.

The main idea of backprop error normalization is to balance the magnitude
of updates in weights (and in threshold) parameters among layers. In the $l$-th layer 
$(N^{(l)} = M^{(l+1)}, n^{(l)} = m^{(l+1)})$, we define the error propagating back through 
the $i$-th active neuron as
\begin{equation}
\label{eq:delta_norm}
  \delta_i^{(l)}=\frac{g_i^{(l)}}{\bar{g}^{(l)}}\sqrt{\frac{M^{(l+1)}}{m^{(l+1)}}}\sum_j^{n^{(l+1)}}w_{ji}^{(l+1)}\delta_j^{(l+1)}, 
\end{equation}
where $g_i^{(l)}=1/V_{th,i}^{(l)}$, $\bar{g}^{(l)}=\sqrt{E \left[ (g_i^{(l)})^2 \right]} \cong \sqrt{\frac{1}{n^{(l)}}\sum_i^{n^{(l)}}(g_i^{(l)})^2}$.
Thus, with (\ref{eq:weight_condition}), the expectation of the squared sum of errors 
(i.e, $E[ \sum_i^{n^{(l)}}{(\delta_i^{(l)})^2} ]$) can be maintained constant through layers.
Although this was confirmed for the case without a WTA circuit, we found 
that it still approximately holds for networks using WTA. 
Weight and threshold parameters 
are updated as: 
\begin{equation}
\label{eq:update}
  \Delta w_{ij}^{(l)}=-\eta_w\sqrt{\frac{N^{(l)}}{m^{(l)}}}\delta_i^{(l)}\hat{x}_j^{(l)}, \quad
  \Delta V_{th,i}^{(l)}=-\eta_{th}\sqrt{\frac{N^{(l)}}{m^{(l)}M^{(l)}}}\delta_i^{(l)}\hat{a}_i^{(l)},
\end{equation}
where $\eta_w$ and $\eta_{th}$ are the learning rates for weight and threshold parameters, 
respectively. We found that the threshold values tend to decrease through the training epochs
due to SGD decreasing the threshold values whenever the target neuron does not fully 
respond to the corresponding input. Small thresholds, however, could lead to exploding 
firing within the network.  Thus, we used smaller learning rates for threshold updates to 
prevent the threshold parameters from decreasing too much. $\hat{x}$ and $\hat{a}$ 
in (\ref{eq:update}) are the effective input and output activities defined as: $\hat{x}_j=x_j$,  
$\hat{a}_i=\gamma a_i - \sigma \sum_{j \neq i}^n{\kappa_{ij}a_j}$.
By using (\ref{eq:update}), at the initial stage of training, 
the amount of updates depends on the expectation of per-synapse activity of active inputs, 
regardless of the number of active synapses or neurons. Thus, we can balance updates 
among layers in deep SNNs. 

\section{Regularization}
\label{regularization}

As in conventional ANNs, regularization techniques such as weight decay during training 
are essential to improve the generalization capability of SNNs. 
Another problem in training SNNs is that because thresholds need to be initialized to large values,
only a few neurons respond to input stimuli and many of them remain silent. This 
is a significant problem, especially in WTA circuits. In this section we introduce weight 
and threshold regularization methods to address these problems. 

\subsection{Weight Regularization}
\label{w_regularization}

Weight decay regularization is used to improve the stability of SNNs as well as their 
generalization capability. 
Specifically, we want to maintain 
the condition in (\ref{eq:weight_condition}). Conventional L2-regularization 
was found to be inadequate for this purpose, because it leads to an initial fast growth, followed by 
a continued decrease of weights. To address this issue, a new method named 
exponential regularization is introduced, which is inspired from max-norm regularization 
\cite{srivastava2014dropout}. The cost function of exponential regularization for neuron $i$ of 
layer $l$ is defined as: 
\begin{equation}
\label{eq:w_reg}
  L_w(l,i)=\frac{1}{2}\lambda e^{\beta \left( \sum_j^{M^{(l)}}(w_{ij}^{(l)})^2 - 1\right)},
\end{equation}
where $\beta$ and $\lambda$ are parameters to control the balance between error 
correction and regularization. 
L2-regularization has a constant rate of decay regardless 
of weight values, whereas max-norm regularization imposes an upper bound of weight 
increase. Exponential regularization is a compromise between the two. The decay 
rate is exponentially proportional to the squared sum of weights. Thus, it strongly prohibits 
the increase of weights like max-norm regularization. Weight parameters are always 
decaying in any range of values to improve the generalization capability as in L2-regularization. 
However, exponential regularization prevents weights from decreasing too much by 
reducing the decay rate. Thus, the magnitude of weights can be easily maintained at a 
certain level. 

\subsection{Threshold Regularization}
\label{th_regularization}

Threshold regularization is used to balance the activities among $N$ neurons receiving 
the same input stimuli. When $N_w$ neurons fire after receiving an input spike, their 
thresholds are increased by $\rho N$. Subsequently, for all $N$ neurons, the threshold 
is decreased by $\rho N_w$. Thus, highly active neurons become less sensitive to 
input stimuli due to the increase of their thresholds. On the other hand, rarely active 
neurons can respond more easily for subsequent stimuli. 
Because the membrane potentials are restricted to the range $[-V_{th}, V_{th}]$,
neurons with smaller thresholds, because of their tight lower bound, tend to be less influenced by negative inputs. 
Threshold regularization actively prevents dead neurons and encourages all neurons 
to equally contribute to the optimization. This kind of regularization has been used for 
competitive learning previously \cite{rumelhart1985feature}. We set a lower bound on thresholds to prevent 
spiking neurons from firing too much due to extremely small threshold values. If the 
threshold of a neuron is supposed to go below the lower bound, then instead of 
decreasing the threshold, all weight values of the neuron are increased by the same 
amount. Threshold regularization was done during the forward propagation in training.

\section{Results and Discussion}
\label{result}

Using the regularization term from (\ref{eq:w_reg}), the objective function for each 
training sample (using batch size = 1) is given by $L=\frac{1}{2}\|a-y\|^2 \ + \sum_{l \in hidden}\sum_i{L_w(l,i)}$
, where $y$ is the label vector and $a$ is the output vector. Each element of $a$ is 
defined as $a_i=\#spike_i/\max_j(\#spike_j)$, where $\#spike_i$ is the number of output 
spikes generated by the $i$-th neuron of the output layer. The output is normalized 
by the maximum value instead of the sum of all outputs. With this scheme, it is not
necessary to use weight regularization for the output layer. 

\begin{table}[t]
  \caption{Values of parameters used in the experiments}
  \label{param_table}
  \centering
  \begin{tabular}{llll}
    \toprule
    Parameters     & Values      & Used In \\
    \midrule
    $\tau_{mp}$  & 20 ms (MNIST), 200 ms (N-MNIST)  & (\ref{eq:lifmp}), (\ref{eq:activity})  \\
    $T_{ref}$        & 1 ms  & (\ref{eq:lifmp})  \\
    $\alpha$       & $3 - 10$    & (\ref{eq:initialization})   \\
    $\eta_{w}$    & $0.002 - 0.004$ & (\ref{eq:update}) \\
    $\eta_{th}$    & $0.1\eta_w$ (SGD), $\eta_w$ (ADAM) & (\ref{eq:update})   \\
    $\beta$         & 10 & (\ref{eq:w_reg})  \\
    $\lambda$    & $0.002 - 0.04$  & (\ref{eq:w_reg}) \\
    $\rho$          & $0.00004 - 0.0002$ & \ref{th_regularization}  \\
    \bottomrule
  \end{tabular}
\end{table}

\begin{figure}[h]
  \centering
  \includegraphics[width=\textwidth]{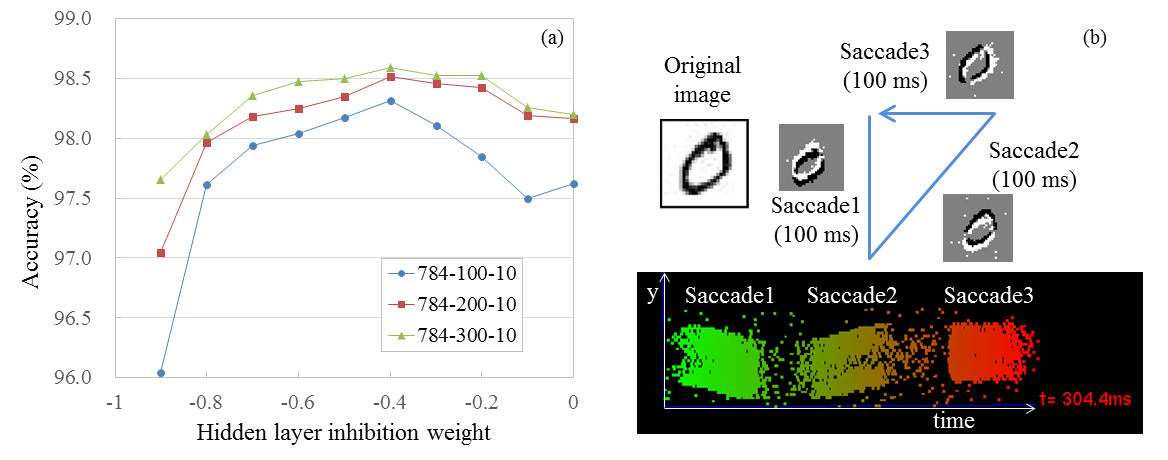}
  \caption{(a) Accuracy vs. strength of lateral inhibition in the hidden layer for PI MNIST. 
     (b) Illustration of the saccades used to generate the N-MNIST dataset and resulting event streams \cite{orchard2015converting}.}
\label{fig1}
\end{figure}

The PI MNIST task was used for performance evaluation \cite{lecun1998gradient}. 
MNIST is a hand written digit classification dataset consisting of 60,000 training samples 
and 10,000 test samples. The permutation-invariant version was chosen to directly 
measure the power of the fully-connected classifier. By randomly permuting the input 
stimuli we prohibit the use of techniques that exploit spatial correlations within inputs, 
such as data augmentation or convolutions to improve performance. An event stream 
is generated from a $28 \times 28$ pixel image of a hand written digit at the 
input layer. The intensity of each pixel defines the event rate of Poisson events. 
We normalized the total event rate to be 5 keps ($\sim$43 eps per non-zero pixel on 
average). The accuracy of the SNN tends to improve as the integration time (i.e. the 
duration of the input stimuli) increases. We used 1 second duration of the input event 
stream during accuracy measurements to obtain stable results. Further increase of integration
time improved the accuracy only marginally ($<0.1\%$). During training, only 50 ms presentations 
per digit were used to reduce the training time. In the initial phase of training deep SNNs, 
neuron activities tend to quickly decrease propagating into higher layers due to non-optimal 
weights and large thresholds. Thus, for the networks with 2 hidden layers (HLs), the 
first epoch was used as an initial training phase by increasing the duration of the input 
stimuli to 200 ms. All 60,000 samples were used for training, and 10,000 samples for testing. 
No validation set or early stopping were used. Learning rate and threshold regularization 
were decayed by $\exp(-1/35)$ every epoch. Typical values for parameters are listed in 
Table \ref{param_table}. We trained and evaluated SNNs with different sized hidden 
layers (784-$N$-10, where $N$ = 100, 200, 300) and varied the strength of lateral inhibitory 
connections in WTA circuits (in the HL and the output layer) to find their optimum value. 
All the networks were initialized with the same weight values and trained for 
150 epochs. The reported accuracy is the average over epochs [131, 150], which reduces the 
fluctuation caused by random spike timing in the input spike stream and training. Figure 
\ref{fig1}(a) shows the accuracy measured by varying the lateral inhibition strength in 
the first HL. The best performance was obtained when  the lateral inhibition was at -0.4, 
regardless of $N$. For the output layer, we found that -1.0 gave  the best result. 
Table \ref{accuracy_table} show the accuracies of various shallow and deep architectures 
in comparison with previous reports. For the deep SNNs with 2 HLs, the first HL and the 
output layer were competing in a WTA circuit. The strength of the lateral inhibition was 
-0.4 and -1.0 for each one as in the case of the SNNs with 1 HL. However, for the second 
HL, the best accuracy was obtained without a WTA circuit, which possibly means that 
the outputs of the first hidden layer cannot be sparsified as much as the original inputs 
without losing information. The best accuracy ($98.64\%$) obtained from the SNN with 
1 HL was better than that of the shallow ANN (i.e. MLP) ($98.4\%$) and matched the previous 
state-of-the-art of deep SNNs \cite{diehl2015fast,hunsberger2015spiking}. We attribute this improvement to the use of WTA circuits 
and the direct optimization on spike signals. The best accuracy of SNN with 2 HLs was 
$98.7\%$ with vanilla SGD. By applying the ADAM learning method ($\beta_1=0.9$, $\beta_2=0.999$, 
$\epsilon=10^{-8}$) \cite{kingma2014adam}, we could further improve the best accuracy up to $98.77\%$, 
which is in the range of ANNs trained with Dropout or DropConnect \cite{srivastava2014dropout,wan2013regularization}. 

To investigate the potential of the proposed method on event stream data, we trained 
simple networks with 1 HL on the N-MNIST dataset, a neuromorphic version 
of MNIST. It was generated by moving a Dynamic Vision Sensor (DVS) \cite{lichtsteiner2008dvs} in 
front of projected images of digits \cite{orchard2015converting}. A 3-phase saccadic movement of the DVS (Figure \ref{fig1}(b)) is responsible
for generating events, and shifts the position of the digit in pixel space. The previous state-of-the-art 
result achieved $95.72\%$ accuracy with a spiking convolutional neural 
network (CNN) \cite{neil2016effective}. Their approach was based on \cite{diehl2015fast}, converting an ANN
to an SNN instead of directly training on spike trains. This lead to a large accuracy drop after conversion ($98.3\% \rightarrow 95.72\%$),
even though the event streams were pre-processed to center the digits.
In this work, however, we work directly on the original uncentered data. For training, 300 consecutive events were picked at 
random positions from each event stream, whereas the full event streams were used for evaluating the test accuracy.
Since the DVS generated two types of event (on-event for intensity increase, off-event for intensity decrease), 
we separated events into two channels based on the event type. 
Table~\ref{accuracy_table} shows that our result of $98.53\%$ with 500 hidden units is the best N-MNIST result with SNNs reported to date.

We have shown that our novel spike-based backpropagation technique for deep SNNs works both on standard benchmarks such as PI MNIST,
but also on N-MNIST, which contains rich spatio-temporal structure in the events generated by a neuromorphic vision sensor.
We improve the previous state-of-the-art of SNNs on both tasks and achieve accuracy levels that match those of conventional deep networks.
Closing this gap makes deep SNNs attractive for tasks with highly redundant information or energy constrained 
applications, due to the benefits of event-based computation, and  advantages of efficient neuromorphic processors \cite{merolla2014million}.
We expect that the proposed technique can precisely capture the statistics 
of spike signals generated from event-based sensors, which is an important advantage over  
previous SNN training methods. Future work will extend our training approach to new architectures, such as CNNs
and recurrent networks.

\begin{table}[t]
  \caption{Comparison of accuracy of different models on PI MNIST 
                without unsupervised pre-training or cost function (except SNN(\cite{oconnor2013real}) and SNN(\cite{hunsberger2015spiking})) 
                and N-MNIST \cite{orchard2015converting}.}
  \label{accuracy_table}
  \centering
  \begin{tabular}{lll}
    \toprule
    Network     & \# units in HLs      & Test accuracy (\%) \\
    \midrule
    ANN (\cite{srivastava2014dropout}, Drop-out)   & 4096-4096  & 98.99  \\
    ANN (\cite{wan2013regularization}, Drop-connect)       & 800-800    & 98.8   \\
    ANN (\cite{goodfellow2013maxout}, maxout)    & 240 $\times$ 5-240 $\times$ 5 & 99.06 \\
    \midrule
    SNN (\cite{oconnor2013real})$ ^{a,b}$   & 500-500  & 94.09   \\
    SNN (\cite{hunsberger2015spiking})$ ^a$  & 500-300 & 98.6 \\
    SNN (\cite{diehl2015fast})    & 1200-1200  & 98.64 \\
    SNN (\cite{oconnor2016deep})    & 200-200  & 97.8 \\
    SNN (SGD, This work)          & 800  & [98.56, 98.64, 98.71]$ ^*$  \\
    SNN (SGD, This work)   & 500-500  & [98.63, 98.70, 98.76]$ ^*$  \\
    SNN (ADAM, This work)   & 300-300  & [98.71, 98.77, 98.88]$ ^*$ \\
    \midrule
    \midrule
    N-MNIST (centered), ANN (\cite{neil2016effective})   & CNN  & 98.3  \\
    N-MNIST (centered), SNN (\cite{neil2016effective})   & CNN  & 95.72  \\
    N-MNIST (uncentered), SNN (This work)  & 500  &  [98.45, 98.53, 98.61]$ ^*$ \\
    \bottomrule
  \end{tabular}
  \\ a: pretraining, b: data augmentation,  *:[min, average, max] values over epochs [181, 200].
\end{table}

\small

\bibliographystyle{abbrv}

\bibliography{arXiv_2016_SNNbackprop}

\end{document}